\documentclass[runningheads,a4paper]{llncs}
\usepackage{hyperref}
\usepackage{cleveref}
\usepackage{multirow}
\usepackage{amssymb}
\usepackage{graphicx}
\usepackage[table,x11names]{xcolor}

\usepackage{url}
\newcommand{\keywords}[1]{\par\addvspace\baselineskip
\noindent\keywordname\enspace\ignorespaces#1}

\begin{document}

\mainmatter  % start of an individual contribution

% first the title is needed
\title{A two-stage 3D Unet framework for multi-class segmentation on full resolution image}
% a short form should be given in case it is too long for the running head
%\titlerunning{Lecture Notes in Computer Science: Authors' Instructions}

% the name(s) of the author(s) follow(s) next
%
% NB: Chinese authors should write their first names(s) in front of
% their surnames. This ensures that the names appear correctly in
% the running heads and the author index.
%
\author{Chengjia Wang $ ^{1,2} $%
\thanks{This work is funded by BHF Centre of Cardiovascular Science and MICCAI 2017 Multi-Modality Whole Heart Segmentation (MM-WHS) challeng.}%
\and Tom MacGillivray$ ^{2} $\and Gillian Macnaught$ ^{1,2} $\and Guang Yang$ ^{3} $\and David Newby$ ^{1,2} $}

%\author{************* $ ^{1,2} $%
%\thanks{This work is funded by *******************************.}%
%\and ****************$ ^{2} $\and *****************$ ^{1,2} $\and **********$ ^{3} $\and ***********$ ^{1,2} $}

% the affiliations are given next; don't give your e-mail address
% unless you accept that it will be published

\institute{$^{1}$BHF Centre for Cadiovascular Science, University of Edinburgh, Edinburgh, UK \url{chengjia.wang@ed.ac.uk} \\
$^{2}$Edinburgh Imaging Facility QMRI, University of Edinburgh, Edinburgh, UK\\
$^{3}$National Heart \& Lung Institute, Imperial College London, London, UK\\
}

%\institute{$^{1}$**************************************************** ************************ \\
%\url{**********************} \\
%$^{2}$*********************************************************
%**************\\
%$^{3}$********************************************************** ***********\\
%}

%
% NB: a more complex sample for affiliations and the mapping to the
% corresponding authors can be found in the file "llncs.dem"
% (search for the string "" where a contribution starts).
% "llncs.dem" accompanies the document class "llncs.cls".
%

\toctitle{Lecture Notes in Computer Science}
\tocauthor{Authors' Instructions}
\maketitle

\begin{abstract}
Deep convolutional neural networks (CNNs) have been intensively used for multi-class segmentation of data from different modalities and achieved state-of-the-art performances. However, a common problem when dealing with large, high resolution 3D data is that the volumes input into the deep CNNs has to be either cropped or downsampled due to limited memory capacity of computing devices. These operations lead to loss of resolution and increment of class imbalance in the input data batches, which can downgrade the performances of segmentation algorithms. Inspired by the architecture of image super-resolution CNN (SRCNN) and self-normalization network (SNN), we developed a two-stage modified Unet framework that simultaneously learns to detect a ROI within the full volume and to classify voxels without losing the original resolution. Experiments on a variety of multi-modal volumes demonstrated that, when trained with a simply weighted dice coefficients and our customized learning procedure, this framework shows better segmentation performances than state-of-the-art Deep CNNs with advanced similarity metrics.
\end{abstract}

\section{Introduction}
Segmenting the whole heart structures from CT and MRI data is a necessary step for pre-precedural planing of cardiovascular diseases. Although it is the most reliable approach, manual segmentation is very labor-intensive and subject to user variability \cite{Pace2015}. High anatomical and signal intensity variations make automatic whole heart segmentation a challenging task. Previous methods that separately segment specific anatomic structure \cite{Petitjean2015,Arrieta2017} are often based on active deformation models. Others perform multi-class segmentation among which atlas-based methods \cite{Zhuang2016} play an important role. Active deformation models can suffer from limited ability to decouple pose variation \cite{Gonzalez-Mora2007}, and the main disadvantage of atlas-based methods is requiring complex procedures to construct the atlas or non-rigid registration \cite{Marsland2003}. Recently, due to the development of deep learning, deep convolutional neural networks (DCNNs) and probabilistic graphic models (PGMs), especially U-net-like models \cite{Ronneberger2015}, have been vastly used for cardiac segmentation and achieved start-of-the-art results. The purpose of this study is to develop a DCNN which can perform multi-class segmentation on full-resolution volumetric CT and MR data with no post-prediction resampling or subvolume-fusion operations. This is necessary due to the loss of information introduced by interpolation and extra complexity of post-processing. 

%Roth2017, Luo2016, Tan2017, Moeskops2016, Mortazi2017, Luo2017}. 

The original U-Net is entirely an 2D architecture. So are most DCNN-based full heart segmentation methods \cite{Wolterink2016} \cite{Moeskops2016}. To process volumetric data, some models takes three perpendicular 2D slices as input and fuse the multi-view information abstraction for 3D segmentation \cite{Mortazi2017,Luo2017}. 3D U-Net-like DCNNs, where the 2D operations were replaced by their 3D counterparts \cite{Cicek2016,Roth2017}, were adopted in different applications. Very limited numbers of works have applied volumetric U-Nets to 3D whole heart CT or MRI data for multi-class segmentation \cite{Yu2017}. Due to limited memory capacity of GPUs, these 3D DCNN methods have to either make predictions on down-sampled volumes, which leads to loss of resolution in the final results, or process subvolumes of the data followed by extra post-processing step to merge the overlapped predictions as in \cite{Yu2017}. Methods that preserve the original data resolution often use a relatively shallow U-Net architecture or have just been tested on low-resolution MR images.

In this paper, we propose a two-stage DCNN framework which is built by concatenating two U-Net-like networks. A new multi-stage learning pipeline was adopted to the training process. This framework with auxiliary outputs segment 3D CT and MR data through dynamic ROI-extraction. Experiments with limited training data have demonstrated that our model outperformed well-trained 3D U-Nets with necessary post-processing steps.

\section{Method} 
\begin{figure}
\centering
\includegraphics[height=6.2cm]{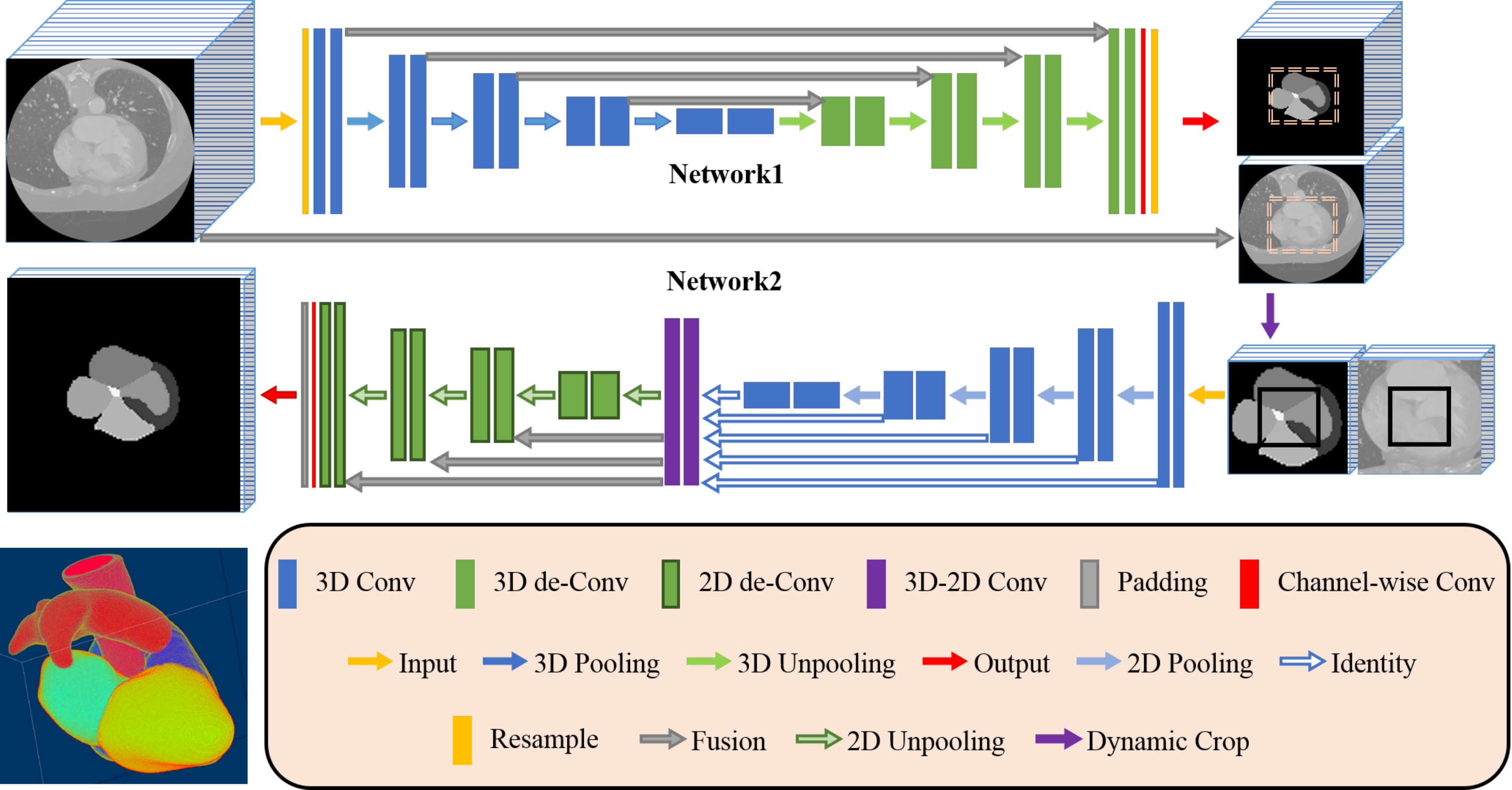}
\caption{The concatenated U-Net architecture proposed in this work. The nonlinear activation and pooling layer within each U-Net block are not shown for demonstration purpose.}
\label{fig:architecture}
\end{figure}
\subsection{DCNN Architecture}
The proposed DCNN model classify all the voxels within an axial slice based on a pre-defined neighborhood of axial slices around it. As shown in Fig. \ref{fig:architecture} the complete model consists of two concatenated modified U-Nets. Generally follow architecture of the original 2D U-Net, the basic block of this model consists of two convolutional layers, each followed  followed by nonlinear activation and a $2 \times 2 \times 2$ pooling layer. Both the contracting (encoding) and the expansive (decoding) paths of the two Unets have 4 basic U-Net blocks \cite{Ronneberger2015}. (Note that the activation and pooling layers within each block are not shown in Fig. \ref{fig:architecture} for clear demonstration.) Each of the two concatenated networks has 23 convolutional layers. The final outputs of both networks are produced by a softmax classification layer.

The first network (\textit{Net1} in Fig. \ref{fig:architecture}) use down-sampled 3D volume to make a coarse prediction of the voxel labels. The produced label volume is then resampled to the original resolution. To capture information from larger effective receptive field, we use slightly dilated $ 5\times5\times5 $ convolutional kernel with zero-padding which preserves shapes of feature maps. In the $ n $th block of the contracting path, the dilation rate of the convolutional kernel is $ 2n $. This pattern is reversed in the expansive path. Each convolutional layer is followed by a rectified linear unit (ReLU), and a dropout layer with a 0.2 dropout rate is attached to each U-Net block. In the test phase, a dynamic-tile layer is introduced between \textit{Net1} and \textit{Net2} to crop out a region-of-interest (ROI) from both the input and output volume of \textit{Net1}. This layer is removed when performing end-to-end training to simplify implementation. 

The architecture of \textit{Net2} is inspired by the deep Super-Resolution Convolutional Neural Network (SRCNN) \cite{Dong2016} with skip connections and recursive units \cite{Kim2016}. The input of this network is a two-channel 4D volume composed by the output of \textit{Net1} and the original data. The convolutional kernel size in the contracting path is $ 3 \times 3 \times 3 $, and $ 5 \times 5 \times 5 $ in the expansive path. Different from $Net1$, the size of the 3D pooling kernels in the contracting path is $ 2 \times 2 \times 1 $ to keep the number of axial slices. A 3D-2D slice-wise convolution block with $ 1\times1\times(K-1) $ convolutional kernels are introduced before the expansive path, where $ K $ is the number of neighboring slices used to label one single axial slice. No zero-paddings are used so that every $K$ input slices will generate one single axial feature map. Furthermore, $ K $ should always be an odd number to prevent generating labels for interpolated slices. The following layers before the output of \textit{Net2} perform 2D convolutions and pooling.

\subsection{Training}
The two U-Net-like DCNNs of the proposed model are flexible enough to be trained either separately or end-to-end with changing sizes of input data. In this study, we combined both approaches into a four-step training procedure. At the beginning, $Net1$ is pre-trained for initial localization of the object. Then the whole framework is trained with different combinations of $Net1$ and $Net2$ loss functions for quick convergence. Details of training batches and loss functions used in different steps are shown in Table \ref{tab:trainsteps}.

\begin{table*}[!t]
\caption{Perposes and loss functions of each step in the training process}
\begin{center}
\begin{tabular}{c@{\quad}l@{\quad}l@{\quad}l@{\quad}l@{\quad}}
\hline\rule{0pt}{12pt}
Step&\multicolumn{1}{c}{Input} & \multicolumn{1}{c}{Purpose}& \multicolumn{1}{c}{Loss}\\
\hline
1 & full volumetric data  &foreground localization& $\mathcal{L}_{ROI}^{1}$\\
2 & partial volumetric data	& coarse multi-class segmentation & $\mathcal{L}_{ROI}^{1} + \mathcal{L}^{1}$\\
3 & partial volumetric data & coarse+fine segmentation & $\mathcal{L}^{1} + \mathcal{L}^{2}$\\
4 & stack of full axial slices & fine multi-class segmenation & $\mathcal{L}^{2}$\\ 
\hline
\end{tabular}
\end{center}
\label{tab:trainsteps}
\end{table*}
\subsubsection{Dice Score}
A commonly used similarity metric for single-class segmentation is soft Dice score. Let $ p_{n,c}^{i} $ denote the probability that a voxel belongs to class $c, c \in \left\{0, \cdots ,C\right\}$, given by the softmax layer of $ Neti $, and $ t_{n,c} \in \left\{ 0, 1 \right\}$ represent the ground truth one-hot label. The soft Dice score can be defined by:
\begin{equation}
\mathcal{S}_{c}^{i} = \frac{2\sum_n^{N_c}t_{n,c}p_{n,c}^{i} + \epsilon}{ \sum_n^{N_c}\left( t_{n,c}+p_{n,c}^{i} \right) + \epsilon}, 
\label{equ:singledicescore}
\end{equation}
where $ N_c $ is number of voxels labeled as class $c$ and $\epsilon$ is a smooth factor. To perform multi-class segmentation, we just define our loss function using weighted Dice scores weighted by voxel counts for simplicity:
\begin{equation}
\mathcal{L}^{i} = 1-\sum\nolimits_c^{C}\frac{S_c^{i}}{N_c}.
\label{equ:diceloss}
\end{equation}
But nothing stops using a more sophisticated loss functions as shown in \cite{Berger2017}. In different steps of the training process, losses of the two nets are combined for different stage targets.

\subsubsection{Foreground Localization}
In this step, \textit{Net1} is trained with full volumetric data to roughly localize the foreground, or a soft ROI, which is the segmented object. Other contents in the data are considered as background. Parameters of $Net2$ is frozen after initialized. The input data is firstly resampled to very coarse resolution, for example, $ 3 \times 3 \times 3 $ as used in our experiments. To encourage localization of foreground, the loss functrion is defined by combining the foreground Dice score with the multi class Dice score. The foreground Dice score $\mathcal{L}_{ROI}^{1}$ computed from $Net1$ output is defined as:
\begin{equation}
\mathcal{S}_{ROI}^{1} = \frac{2 \sum\nolimits^{N_0}_{n} \left( 1-t_{n,0} \right) \left( 1- p_{n, 0}^{1} \right) + \epsilon }{\sum\nolimits^{N_{0}}_{n}\left( 2-t_{n,0} - p_{n,0}^{1}  \right) + \epsilon},
\label{equ:roiscore}
\end{equation}
where $N_{0}$ is the number of the background points as the background is defined as class $0$. The corresponded foreground loss is:
\begin{equation}
\mathcal{L}_{ROI}^{1} = 1- \frac{S_{ROI}^{1}}{N_{0}}.
\label{equ:roiloss}
\end{equation}
We use reversed label to calculate foreground score rather than the Dice score of background to reduce the imballance introduced by large background. $Net1$ is trained to minimize the loss $\mathcal{L}_{ROI}^{1} $ for quickly specify the foreground of the object.

\subsubsection{Multi-class Segmentation}
After pre-training with $\mathcal{L}_{ROI}^{1}$, $ \mathcal{L}^{1} + \mathcal{L}_{ROI}^{1} $ is used as the loss for coarse multi-class segmentation in the second step, where $\mathcal{L}^{1}$ is Dice loss defined by equation \ref{equ:diceloss}. In this step, $ Net1 $ is trained using subvolumes of the data. Dimensions of the data are varied in different training batches as an augmentation strategy. In the third step, the whole framework (both $Net1$ and $Net2$) is trained end-to-end with the loss $\mathcal{L}^{1} + \mathcal{L}^{2}$ to evolve both coarse 3D segmentation and the fine-level axial slice segmentation. As both networks are fully convolutional, the sampling strategy of the input data keep the same with step 2. In the final step, inputs of the framework are subvolumes, each consist of $K$ complete axial slices. The output of $Net2$ is the segmentation of the $\frac{K+1}{2}$th slice of a input subvolume. In this step, the parameters of $Net1$ are frozen, and $Net2$ is finetuned using the loss $ \mathcal{L}^{2} $.

\subsection{Implementation Details}
Because the framework is mostly trained with subvolumes of the 3D data except the first step, we use a hierarchical sampling strategy similar with \cite{Girshick2015}. Each batches are generated from a small number of data. Dealing with highly imbalanced data, we first select the class that the central voxel of the sampled subvolume from a uniform distribution. Once the label of the central voxel is fixed, the subvolume is generated by randomly pick its centre from all voxels labeled as the selected class. In this way, the probabilities that the central voxel belongs to any of the classes should be $\frac{1}{C+1}$.
For optimization, we use Adam optimizer with initial learning rate $ 0.0001 $. 

We use only 1 full volume in each batch for initial training in the first step. For partial volumetric data in step 2 and 3, the size of each dimension of the data is randomly pick from $\left\{ 64, 128, 256 \right\}$. In the final training step, we set $ K=9 $, which means $Net2$ take a subvolume that contains 9 axial slices and predict the labels for the 5th slice of this subvolume. Tensorflow code was adopted on Microsoft Azure virtual machine with one nVidia Tesla K40 GPU, which has 12G memory. Input data augmentation include random rotation, translation, sheering, scaling, flipping and elastic deformations.

\section{Experiments and Data}
The MICCAI 2017 Multi-Modality Whole Heart Segmentation (MM-WHS) challenge recently benchmarks existing whole heart segmentation algorithms. For training purpose, the challenge provides 40 volumes (20 cardiac CT and 20 cardiac MR) in the real clinical environment. The data were acquired with different scanners, which leads to varying voxel sizes, resolutions and imaging qualities. An extra 80 testing images are available from the challenge for one-shot validation. In this dataset, anatomical structures which are manually delineated include, the left ventricle blood cavity (LV), the myocardium of the left ventricle (Myo), the right ventricle blood cavity (RV), the left atrium blood cavity (LA), the right atrium blood cavity (RA), the ascending aorta (AA) and the pulmonary artery (PA).

One may argue that the proposed framework can be trained directly using the final training step. To demonstrate their effectivenes, we trained our model by omitting one of the first three steps, and visually assessed the segmentation results which can be found in the next section. The four training steps don't have to be kept going until converged except the final step. In this step-wise experiment, all results were obtained with 200 epochs in each step, and each epoch includes 16 iterations of backpropogation. The whole training process contains 12800 iterations in total. 

Besides qualitatively evaluating the visualized segmentation, for each modality, we use 15 volumes for 3-fold cross-validation training, and 5 volumes for validation. To compare our framework with state-of-the-art U-Net-based models, we trained two 3D U-Nets for each modality which predict on data resampled to resolution of $ 2\times2\times2 mm^{3} $. Then the output volumes are resampled to the original resolution using 2nd order BSpline interpolation. Intensities of all images are rescaled to $ \left[ -1, 1 \right] $ with no further preprocessing. Three metrics were used to assess segmentation quality for each class: binary Dice score (Dice), binary Jaccard index (Jaccard). After step-wise experiments, we retrain the networks end-to-end from scratch and submit the results to MM-WHS challenge for further evaluation using their test data.

\section{Results}
\begin{figure}[!h]
\centering
\includegraphics[width=\textwidth]{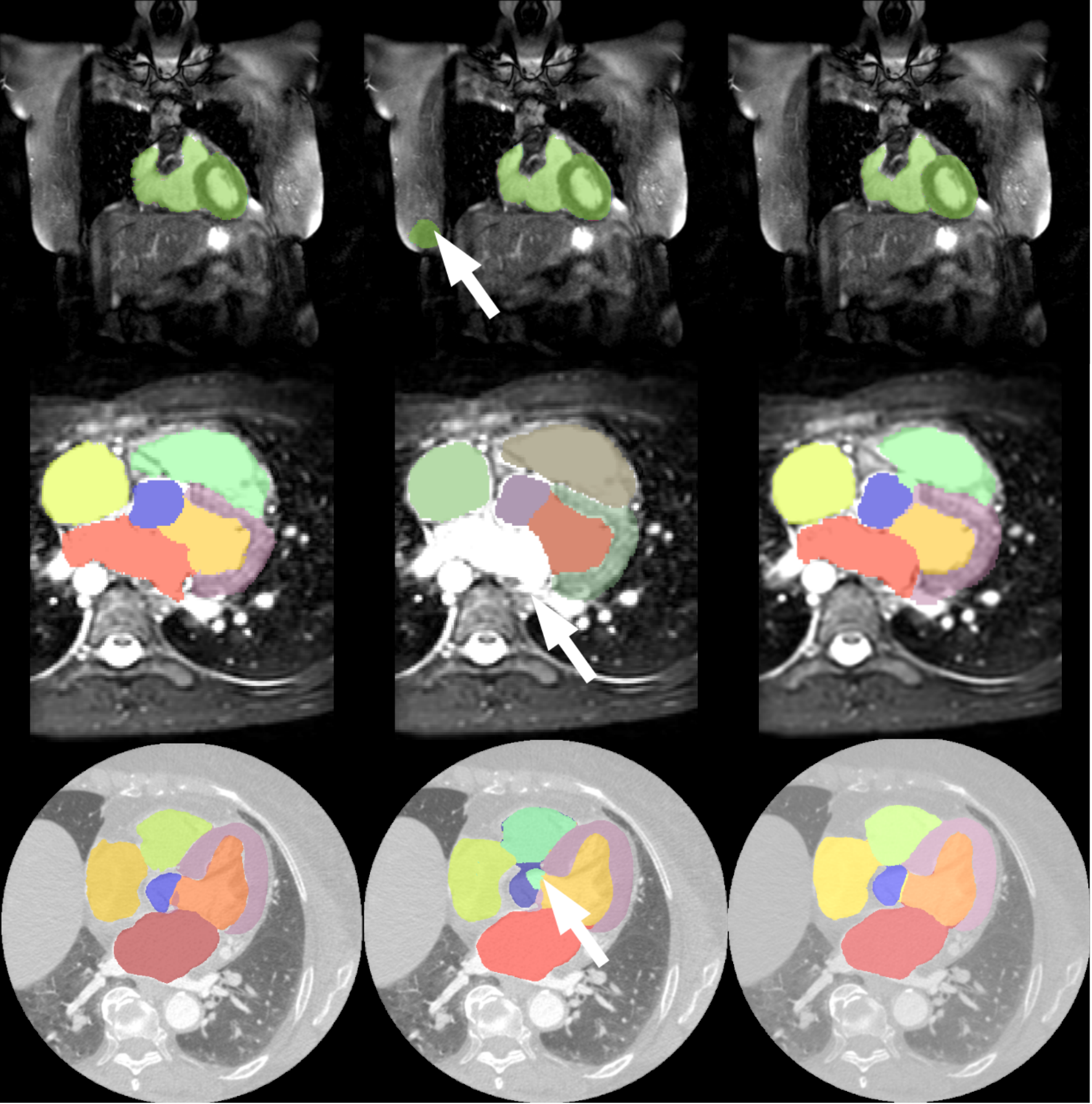}
\caption{Visualization of segmentation results overlapped with the original data: Ground-truth segmentations are shown on the left; the middle column shows the results obtained by omitting the first, second and third training step (from top to bottom); on the right is the results obtained with the proposed training process.}
\label{fig:vis}
\end{figure}

Examples of viusalized segmentation reulsts are shown in Fig. \ref{fig:vis}. Omitting the foreground localization step in the first training step may lead to misclassification of the background voxels, as shown in the top row of Fig. \ref{fig:vis}. The middle row shows that without the coarse segmentation (second step) the model failed to label left atrium, and produced inhomogeneious segmentation for aorta when skipping the joint training of $Net1$ and $Net2$ (step three).

Table \ref{tab:CT_cross} and Table \ref{tab:MR_cross} show the binary Dice and Jaccard scores for all assessed structures obtained by $Net1$ and $Net2$ from the proposed framework, compared to individually trained U-Nets. $Net2$ produced highest Dice and Jaccard scores for all segmented structures in CT data, $ Net1 $ gave better results than the individual U-Net trained on the same downsampled data. As the MR data have relatively lower resolution, the volume size  changed less after resampling. $ Net2 $ still produced better segmentation accuracy except for RV and AA. $ Net1 $ gave better segmentations for 4 out of all 7 classes. 

\begin{table*}[!t]
\caption{Comparison of CT segmentation results obtained by 3D U-Net, and our proposed $Net1$ and $Net2$.}
\begin{center}
\begin{tabular}{lcccccccc}
\hline
&\multicolumn{1}{c}{Metrics} & \multicolumn{1}{c}{LV}& \multicolumn{1}{c}{Myo}&\multicolumn{1}{c}{RV}&\multicolumn{1}{c}{LA}&\multicolumn{1}{c}{RA}&\multicolumn{1}{c}{AA}&\multicolumn{1}{c}{PA}\\
\hline
\hline
\multirow{2}{*}{\textbf{N3D U-Net}}&Dice & 0.6451 & 0.8301 & 0.7873 & 0.7768 & 0.6784 & 0.8306& 0.7123 \\
\cline{2-9}
&Jaccard & 0.4889 & 0.7126 & 0.6572 & 0.6397 & 0.5217 & 0.7143 & 0.5560 \\
\hline
\multirow{2}{*}{\textbf{Net1}}&Dice& 0.6774& 0.8107& 0.8136 & 0.8118 & 0.7997 & 0.8889 & 0.8086 \\
\cline{2-9}
&Jaccard & 0.5399 & 0.6979 & 0.6977 & 0.6908 & 0.6717& 0.8030 & 0.6802 \\
\hline
\multirow{2}{*}{\textbf{Net2}}& Dice & \textbf{0.8374} & \textbf{0.8588} & \textbf{0.8600} & \textbf{0.8613} & \textbf{0.8620} & \textbf{0.9176} & \textbf{0.8846} \\
\cline{2-9}
& Jaccard & \textbf{0.7823} & \textbf{0.8210} & \textbf{0.8233} & \textbf{0.8256} & \textbf{0.8268} & \textbf{0.8534} & \textbf{0.7959} \\
\hline
\end{tabular}
\end{center}
\label{tab:CT_cross}
\end{table*}

\begin{table*}[!t]
\caption{Comparison of MR segmentation results obtained by 3D U-Net, and our proposed $Net1$ and $Net2$.}
\begin{center}
\begin{tabular}{lcccccccc}
\hline
&\multicolumn{1}{c}{Metrics} & \multicolumn{1}{c}{LV}& \multicolumn{1}{c}{Myo}&\multicolumn{1}{c}{RV}&\multicolumn{1}{c}{LA}&\multicolumn{1}{c}{RA}&\multicolumn{1}{c}{AA}&\multicolumn{1}{c}{PA}\\
\hline
\hline
\multirow{2}{*}{\textbf{N3D U-Net}}&Dice & 0.8296 & 0.9141 & \textbf{0.9173} & 0.8946 & 0.8792 & \textbf{0.9202} & 0.8847 \\
\cline{2-9}
&Jaccard & 0.7106 & 0.8419 & \textbf{0.8479} & 0.8107 & 0.7894 & \textbf{0.8526} & 0.7938 \\
\hline
\multirow{2}{*}{\textbf{Net1}}&Dice& 0.8811& 0.9367& 0.9131 & 0.9334 & 0.8572 & 0.8750 & 0.9204 \\
\cline{2-9}
&Jaccard & 0.7877 & 0.8813 & 0.8430 & 0.8757 & 0.7694 & 0.7833 & 0.8528 \\
\hline
\multirow{2}{*}{\textbf{Net2}}& Dice & \textbf{0.8813} & \textbf{0.9377} & 0.9125 & \textbf{0.9338} & \textbf{0.9220} & 0.8758 & \textbf{0.9210} \\
\cline{2-9}
&Jaccard & \textbf{0.7879} & \textbf{0.8829} & 0.8422 & \textbf{0.8764} & \textbf{0.8568} & 0.7847 & \textbf{0.8539}  \\
\hline
\end{tabular}
\end{center}
\label{tab:MR_cross}
\end{table*}

The quantitative validation results of $Net2$ obtained using the test dataset of MM-WHS competition are shown in Table \ref{tab:WH_test}. The proposed framework achieved obviously higher accreacy for LA and AA. This is a sign of premature termination of training, although the framework still obtained much better results than the baseline algorithm of the competition. For MR data, the average Dice score of $Net2$ is 0.8323, which is comparable to the winner of the competition. However, our model was only trained 12800 iterations which is only 1/5 of the winner model. Notice that the purpose of this study is to generate the model gave better performance than state-of-the-art U-Net when segmenting high resolution data. This has been shown in the experiment described above.

\begin{table*}[!t]
\caption{Quantative validation results obtained using MM-WHS test data, evaluation metrics include: average of Dice score (Dice), average of Jaccard score (Jaccard), and Average surface distance (ASD).}
\begin{center}
\begin{tabular}{cccccccccc}
\hline
\multicolumn{1}{c}{Modality} & \multicolumn{1}{c}{Metrics} & \multicolumn{1}{c}{LV}& \multicolumn{1}{c}{Myo}&\multicolumn{1}{c}{RV}&\multicolumn{1}{c}{LA}&\multicolumn{1}{c}{RA}&\multicolumn{1}{c}{AA}&\multicolumn{1}{c}{PA}&\multicolumn{1}{c}{WH}\\
\hline
\hline
\multirow{3}{*}{\textbf{CT}} & Dice & 0.7995 & 0.7293 & 0.7857 & 0.9044 & 0.7936 & 0.8735 & 0.6482 & 0.8060\\\cline{2-10}
& Jaccard & 0.6999 & 0.6091 & 0.6841 & 0.8285 & 0.6906 & 0.8113 & 0.5169 & 0.6970 \\
\cline{2-10}
& ASD& 4.4067& 5.4854& 4.8816 & 1.3978 & 4.1707 & 3.7898 & 6.0041 & 4.1971 \\
\hline
\multirow{3}{*}{\textbf{MR}} & Dice & 0.8632 & 0.7443 & 0.8485 & 0.8524 & 0.8396 & 0.8236 & 0.7876 & 0.8323\\
\cline{2-10}
& Jaccard & 0.7693 & 0.6049 & 0.7469 & 0.7483 & 0.7404 & 0.7095 & 0.6657 & 0.7201 \\
\cline{2-9}
& ASD& 1.9916 & 2.3106 & 1.8925 & 1.7081 & 2.7566 & 4.2610 & 2.9296 & 2.4718 \\
\hline
\end{tabular}
\end{center}
\label{tab:WH_test}
\end{table*}

\section{Conclusion and Discussion}
In this paper, we described a two-stage U-Net-like framework for multi-class segmentation. Unlike other U-Net based 3D data segmentation DCNN, the proposed method can directly make prediction for data with original resolution due to its SRCNN-inspired architecture. A novel 4-step training procedure were applied to the framework. Validated using data from MM-WHS2017 competition, it produced more accurate multi-class segmentation results than state-of-the-art U-Net. With much less training iterations and without any further post-processing, our method achieved segmentation accuracies comparable to the winner of MM-WHS2017 competition. 

%\begin{thebibliography}{4}
%
%\bibitem{jour} Smith, T.F., Waterman, M.S.: Identification of Common Molecular
%Subsequences. J. Mol. Biol. 147, 195--197 (1981)
%
%\bibitem{lncschap} May, P., Ehrlich, H.C., Steinke, T.: ZIB Structure Prediction Pipeline:
%Composing a Complex Biological Workflow through Web Services. In: Nagel,
%W.E., Walter, W.V., Lehner, W. (eds.) Euro-Par 2006. LNCS, vol. 4128,
%pp. 1148--1158. Springer, Heidelberg (2006)
%
%\bibitem{book} Foster, I., Kesselman, C.: The Grid: Blueprint for a New Computing
%Infrastructure. Morgan Kaufmann, San Francisco (1999)
%
%\bibitem{proceeding1} Czajkowski, K., Fitzgerald, S., Foster, I., Kesselman, C.: Grid
%Information Services for Distributed Resource Sharing. In: 10th IEEE
%International Symposium on High Performance Distributed Computing, pp.
%181--184. IEEE Press, New York (2001)
%
%\bibitem{proceeding2} Foster, I., Kesselman, C., Nick, J., Tuecke, S.: The Physiology of the
%Grid: an Open Grid Services Architecture for Distributed Systems
%Integration. Technical report, Global Grid Forum (2002)
%
%\bibitem{url} National Center for Biotechnology Information, \url{http://www.ncbi.nlm.nih.gov}
%
%\end{thebibliography}

\bibliographystyle{splncs}
\bibliography{ref}

\end{document}